\documentclass[sigconf]{acmart}

\setcopyright{rightsretained}

\acmConference[SS4HRI '24]{Workshop on Social Signal Modelling}{March 11,
  2024}{Boulder, CO, USA}
\acmBooktitle{SS4HRI: Workshop on Social Signal Modelling, March 11, 2024, Boulder, CO, USA}
\acmDOI{}
\acmISBN{}
\usepackage{multirow}
\usepackage{rotating}

\begin{document}

\title{SSUP-HRI: Social Signaling in Urban Public \\Human-Robot Interaction dataset}

\author{Fanjun Bu}
\affiliation{%
  \institution{Cornell University, Cornell Tech}
  \city{New York}
  \country{USA}}
\email{fb266@cornell.edu}

\author{Wendy Ju}
\affiliation{%
  \institution{Cornell University, Cornell Tech}
  \city{New York}
  \country{USA}}
\email{wendyju@cornell.edu}

\renewcommand{\shortauthors}{Bu and Ju}

\begin{abstract}
  This paper introduces our dataset featuring human-robot interactions (HRI) in urban public environments.\footnote{Dataset information is available at \url{https://github.com/FAR-Lab/SSUP-HRI}.} This dataset is rich with social signals that we believe can be modeled to help understand naturalistic human-robot interaction.  Our dataset currently comprises approximately 15 hours of video footage recorded from the robots' perspectives, within which we annotated a total of 274 observable interactions featuring a wide range of naturalistic human-robot interactions. The data was collected by two mobile trash barrel robots deployed in Astor Place, New York City, over the course of a week. We invite the HRI community to access and utilize our dataset. To the best of our knowledge, this is the first dataset showcasing robot deployments in a complete public, non-controlled setting involving urban residents.
\end{abstract}



\begin{teaserfigure}
  \includegraphics[width=\textwidth]{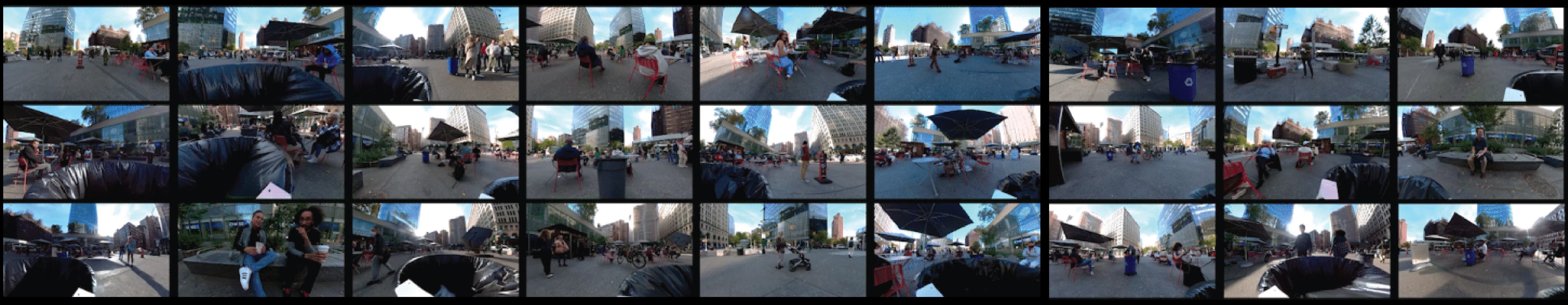}
  \caption{Our dataset features hundreds of human-robot interactions in public spaces from the robots' perspective, which showcase rich social signals in a variety of contexts.}
  \label{fig:teaser}
\end{teaserfigure}


\maketitle

\section{Introduction}
While mobile robots are not yet a common feature in public spaces, the emergence of autonomous agents such as self-driving vehicles and unmanned food delivery robots is on the horizon. It will be important to understand how everyday people who encounter these robots might interact with these robots; as part of this, the social signals that they perform in and around the robots will be important for the robots to pick up on and respond to. 
It is also important to understand how people perceive the movement of the robot itself as a form of social signaling. People inherently attribute social meaning to movement, even of the most basic of shapes \cite{heider1944experimental, smith2005development}. Robots, even those not designed for verbal or social interaction, need to be adept in understanding the signals people make at them, and making sure they respond appropriately in kind. 
This understanding is crucial to ensure they navigate public spaces safely and respectfully, avoiding causing inconvenience or offense to passersby. Moreover, it is imperative for mobile robots to align their movements with the existing dynamics, norms, and expectations of the public spaces they inhabit.

In this paper, we present a unique dataset that captures interactions between people and two mobile trash barrel robots in a bustling public plaza in New York City. These robots are limited to motion as their sole mode of interaction, and are controlled in a Wizard-Of-Oz (WoZ) method to make sure their movements align with human expectations in the situated context. Our dataset will contribute to the social science understanding of public human-robot interactions and assist in the modeling of social signaling in human-robot interactions in the wild.

\section{Related Works}

\subsection{HRI in Public Spaces}
Previous research on human-robot interactions in semi-public settings like hotels and hospitals has provided specific insights related to particular sites and tasks \cite{pan2013direct, pan2015reaction, diaz2015evaluating, onyeulo2020makes, anvari2020modelling}. Broader studies have concentrated on modeling and predicting human movement to ensure robots can navigate without causing obstructions \cite{taylor2020robot,adeli2021tripod, coscia2018long, alahi2017learning}. Furthermore, significant work in social contexts has delved into proxemic behaviors, developing models to determine the ideal distance between robots and humans during interactions to maintain comfort~\cite{mumm2011human, henkel2012proxemic, henkel2012towards}. However, models for cooperative interaction in public also need to account for the social factors that influence whether people want to interact with a robot or not, such as \citet{koile2003activity}'s work on "activity zones" in homes, which highlights spatial influences on interactions. Public interaction is similar, but far more complex, with subtle factors like bench arrangement and sunlight direction significantly influencing behavior patterns and social dynamics \cite{whyte1980social}. 

\subsection{Datasets for HRI}
Datasets focusing on physical interactions between humans and robots are scarce because collecting real-time robotics data is inefficient. Consequently, many datasets are centered around human-human interaction data, with the aim for researchers to use the data to create algorithms applicable to robots \cite{kshirsagar2023dataset, carfi2019multi, ondras2023human}. \citet{joo2019towards} released a dataset of nonverbal signals among people in a triadic social interaction scenario, with the hope of teaching machines nonverbal communication skills.  
In efforts to incorporate robots into data collection, the Wizard-of-Oz method is often employed, wherein robots are remotely controlled by unseen researchers to simulate authentic behaviors \cite{WoZ}. For example, \citet{tian2023crafting} collected a human-robot collaboration dataset featuring handovers with a researcher teleoperating the robot. The MHHRI dataset features triadic interactions between two participants and a robot to study personality and engagement \cite{celiktutan2017multimodal}. The robot's dialogue and expressions are also managed through Wizard-of-Oz. Given the complex and context-sensitive nature of human-robot interactions, the Wizard-of-Oz approach is frequently utilized as a reliable method for capturing genuine interaction data.

Different from the aforementioned datasets, which were all collected in controlled lab settings, our dataset features social signaling in a naturalistic environment without explicitly engineered tasks. Here, the signals from passersby are unprompted. Though the range of activities and factors in the dataset are noisier, models built from the data might be anticipated to be more ecologically valid.

\section{Data Collection}
In this section, we detailed our study deployment and data collection process. All deployments took place in a public space in New York City and complied with local regulations. Our study was approved by Cornell Tech's IRB protocol \#IRB014571. 

\subsection{Robot Deployments}
We deployed two trash barrel robots, one for recycling, and the other for landfill, in Astor Place, New York City. The robots roved the public square and collected trash from people passing by or sitting at the chairs and tables set out in the area. We operated the robots for 2 hours in the early afternoons for 5 days when weather permitted. The hours coincided with high traffic and high occupancy moments in the day. 

Methodologically, we used a semi-supervised Wizard-of-Oz protocol \cite{WoZ, JD_WoZ}. Trained researchers piloted the robot, with the goal of engaging the users as a service robot would. The Wizard-of-Oz method allows us to have the robots respond to social signals in ways that the users were likely to expect; the decisions of the robot operators were as much part of the study as the reactions of the public. For safety reasons, the operators always maintained a line of sight to the robots they were controlling; however, they obscured their role by hiding the controllers or wearing sunglasses so it was less obvious where they were looking. Following people's interactions with the robot, we had another researcher approach to interview willing individuals for their feedback and thoughts on the interaction.

During the deployment, the wizards were encouraged to keep two robots physically close to each other. They were also encouraged to communicate with each other and coordinate the robots' motions. In rare cases where one robot got stuck, the other robot was expected to "bump" the robot in need, instead of direct human intervention.

\subsection{Collection Site}
 To make sure we were collecting naturalistic social signals, we looked for public spaces that were intensively used by residents daily, where the use case of having a mobile robot in the space seemed plausible. Other site selection criteria included foot traffic, availability of public seating, business around the plaza, and ease of research activity operation.

 This deployment eventually took place in Astor Place, a public plaza located in the heart of Manhattan, New York City. This public plaza was a trapezium-shaped traffic island with multiple tables and chairs laid out for people to socialize. Originally we intended the robots to be robot chairs and tables, but in our meetings with the business improvement district that operates the plaza, it was determined that robot trash barrels, like those previously deployed by Yang et al. at Stanford \cite{yang2015experiences}, would fit in better and gain more interaction. The arrangements of the seating were not fixed day to day. As a result, the wizards sat at different locations for each deployment. 

\begin{figure}[h]
  \centering
\includegraphics[width=0.5\textwidth]{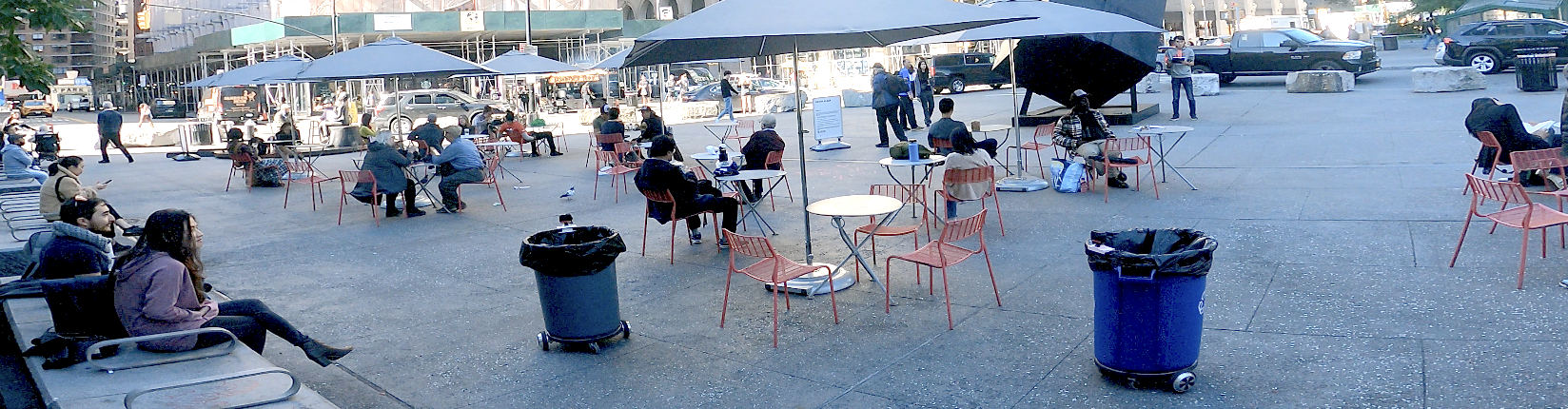}
  \caption{A bird-eye view of our deployment space.}
\end{figure}
 
\subsection{Recording Apparatus}
Each trash barrel robot is a simple mobile platform. Each robot primarily interacts with passersby through its movement. This minimalism helps focus our study and subsequent model on presence and movement, without the confounds that other sounds or visual elements would bring. On the other hand, such design choice also constrains the range of people's signaling towards the robots, which also largely depends on motion and gesture instead of spoken instructions. 

To capture the social signals around them, each trash barrel robot is equipped with an Insta360 One R camera\footnote{\url{https://www.insta360.com/product/insta360-oner_twin-edition}} with built-in microphones and an external IMU to record the surrounding and robot motions. The robots are powered by recycled hoverboards; the robust metal chassis provides the structural integrity needed to navigate outdoor urban terrain. Computationally, each robot currently runs on a Raspberry Pi 4 running the Ubuntu 20.04 server and Robot Operating System (ROS 1 Noetic) \cite{quigley2009ros}.

\begin{figure}[h]
  \centering
\includegraphics[width=.3\textwidth]{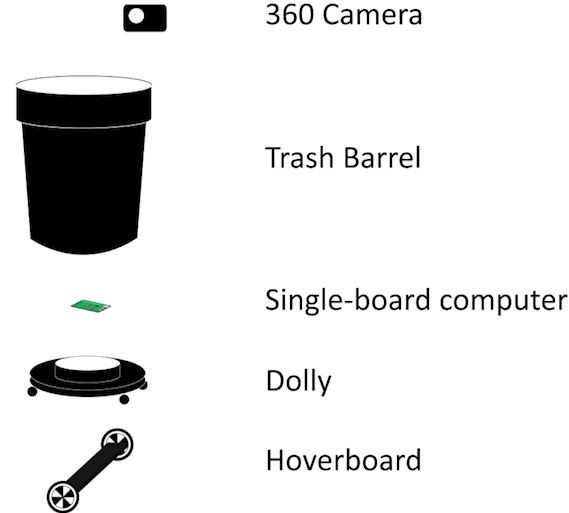}
  \caption{Exploded diagram of robot.}
\end{figure}

\subsection{Data Annotation}
To isolate the social signals in the dataset, two researchers from our research group annotated the dataset and labeled every explicit or implicit interaction moment. Between two of the researchers, one of them was involved in the data collection process, where he was on-site for each deployment. The other researcher was not involved during the collection process. The two researchers labeled the dataset individually and then consolidated a spreadsheet of all interaction moments. The spreadsheet will also be publicly available. Based on our interests, our annotation focused on explicit interactions between plaza users and trash barrel robots. For every interaction, each annotator labeled time, which robot was involved in the interaction, and a summary of the scenario. 

\section{Dataset Overview}
Overall, we observed 274 interactions between the plaza users and trash barrel robots, ranging from normal disposal of trash to people helping the robot to get unstuck. Each interaction consists of unique exchanges of social signals between the users and the robots.

Within the 274 observed interactions, 118 interactions feature people and trash barrel robots negotiated to dispose of trash, for which we provide more detailed annotations focusing on behavioral signaling. Among trash disposal interactions, people interacted with both robots equally (48 interactions with the recycling robot, 54 interactions with the landfill robot, and 16 interactions with both). Over half (\textit{N} = 68) of the interactions feature negotiations between a single user and the robots, while party sizes range from 1 to 5 people. Even though people use different signals to initiate and terminate interactions with the robots, we are able to categorize them into the categories shown in table \ref{signaling}. Lastly, 18 individuals or groups try to have a conversation with the robot during the interaction. 

\begin{table}[htb]
\centering
\begin{tabular}{|l|c|c|}
\hline
\multirow{10}{*}{Initiation}   & Signal Type & Count              \\
            \hline
                               & Passive: Robot approached  & 23   \\
                               & Waving without objects     & 7    \\
                               & Waving/holding objects     & 47   \\
                               & Walking towards robots     & 37   \\
                               & Standing up                & 4    \\
            \hline
\multirow{10}{*}{Termination} 
                               & Walking away                & 47    \\
                               & Verbal                      & 8     \\
                               & Turning around              & 4     \\
                               & Sitting down                & 6     \\
                               & Hand Gesture                & 16    \\
                               & Passive: Robots drove away  & 11    \\
                               & Looking away                & 25    \\
            \hline
\end{tabular}
\caption{Different signals people use to initiate and terminate interactions.}
\label{signaling}
\end{table}

\section{Use Cases}
We anticipate our dataset to significantly impact social science understandings of human-robot interaction in public spaces and aid in advancing social intelligence in machine learning models. Methodologically, we believe our dataset enables bridging qualitative and quantitative practices in the field of human-robot interaction. Pragmatically, our dataset can help establish guidelines to improve the quality of interaction provided by social robots in public.

\subsection{Qualitative Analysis}
Ethnographic researchers may find our video clips valuable and can discover insights generalizable to other applications. Our dataset features naturalistic exchanges of social signals in the context of social navigation and service robotics. Furthermore, interview transcripts can be used for thematic analysis to understand the mental models and thought processes people go through when encountering robots in uncontrolled environments and deciding ways of signaling. Furthermore, the narration and reflection of the encounter depict how users interpret social signals presented by the robots. Our dataset touches upon topics including, but not limited to, people's sensemaking processes, assumptions, expectations, interaction sequences, and cultural influences on human-robot interaction.

\begin{figure*}[h]
    \centering
    \includegraphics[width=\textwidth]{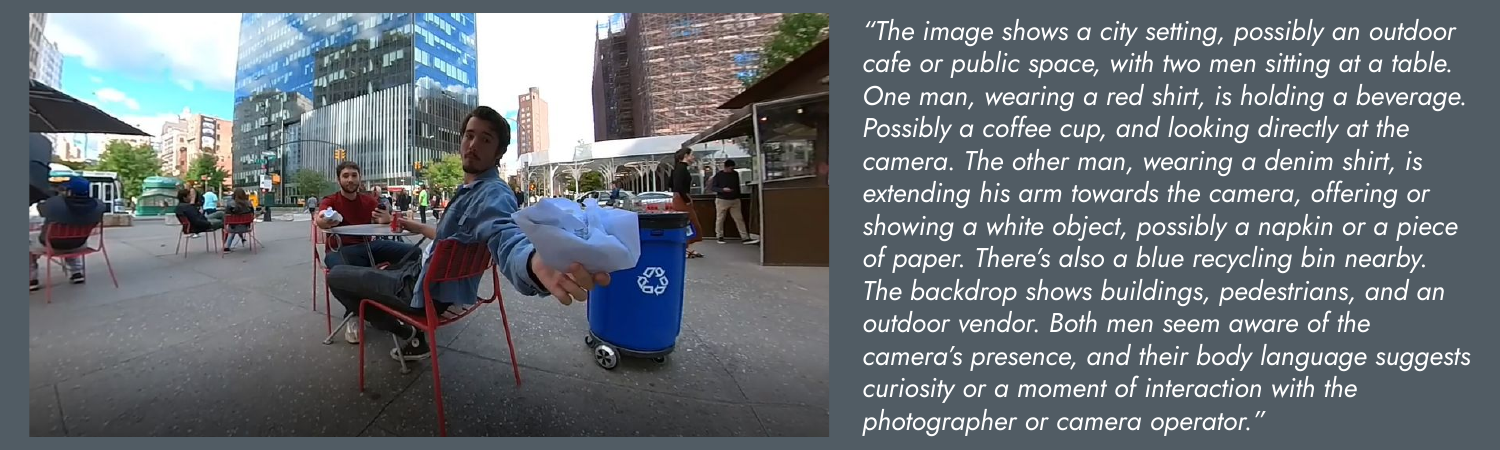}
    \caption{LLM generated captions with inputs from our field study.}
\end{figure*}

\subsection{Quantitative Analysis}
Our dataset also opens room for researchers to perform quantitative analysis to provide numerical support for qualitative analysis. 
From the data, statistics on activities such as people's initiation and termination actions, count party sizes and physical formation, characterize stances (sitting, standing, or walking), and interaction duration are easily observable. Additionally, researchers can profile the robots' actions, based on metrics like approach angles, speed, and distance covered. Our dataset also offers a new playground to analyze existing theoretical models, such as the social force model and F-Formation \cite{kendon1990spatial, cristani2011social, helbing1995social}.

\subsection{Machine Learning Test Bed}
While our dataset volume is not yet large enough to train a machine learning algorithm from scratch, it provides a test bed for computer scientists to test out their algorithms with realistic real-world data. It is possible both to pick out the social signals generated by passersby who want a robot to approach or stay away and also to model what motions and signals the robots are giving off that people are responding to. Furthermore, given the recent progress in the development of foundational models, we are optimistic that our dataset will play a role in establishing a foundation model for interactive mobile robots navigating social environments in urban public spaces.

\section{Future Direction}
This project marks the initial phase of establishing a dataset focused on social interactions between humans and mobile robots within urban environments. Our objective is to extend the deployment of these robots to numerous public plazas, with the ultimate aim of covering all five boroughs of New York City. We intend to expand and enrich our dataset as we continually gather new data.

\section{Conclusion}
We present SSUP-HRI, a dataset that features human-robot interaction in public social spaces. The dataset is exploratory in nature, where both the users and the wizards controlling the robots are trying to decipher the social signals presented by each other. Since the data were collected in unconstrained public space, we observed parties of various sizes negotiated with the robots for various purposes under different contexts. Therefore, we believe our dataset will be beneficial to both researchers and practitioners in the understanding of social signals in the context of human-robot interaction in urban spaces.

\begin{acks}
We thank the Village Alliance for permission to run
the study at Astor Place. This work was supported by research
funding from Tata Consulting Services. This
study would not have been possible without the wizards behind the scenes:
Rei Lee, Saki Suzuki, Ricardo Gonzalez, Stacey Li, Maria Teresa
Parreira, and Jorge Pardo Gaytan.
\end{acks}

\bibliographystyle{ACM-Reference-Format}
\bibliography{dataset}

\end{document}